\documentclass[akbc,twoside,11pt,lettersize]{article}
\newcommand{\method}{XREF}
\usepackage{akbc}
\usepackage{graphicx}
\usepackage{booktabs}
\usepackage{CJKutf8}
\usepackage{amssymb}
\usepackage{amsmath}
\usepackage{subfig}
\usepackage{color,colortbl}
\usepackage{multirow}
\usepackage{bm}
\usepackage{wrapfig}

\definecolor{Gray}{gray}{0.9}

\akbcheading
\ShortHeadings{Xref: Entity Linking for Chinese News Comments}{}
\finalcopy % Uncomment for camera-ready version, but NOT for submission.

\begin{document}

\title{\textsc{XREF}: Entity Linking for Chinese News Comments \\with Supplementary Article Reference}

\author{\name Xinyu Hua \email hua.x@husky.neu.edu \\
        \addr
        Northeastern University \\
        Boston, MA 02115
        \AND
        \name Lei Li \email lileilab@bytedance.com \\
        \addr ByteDance AI Lab, \\
        Beijing, China 
        \AND
        \name Lifeng Hua \email issac.hlf@alibaba-inc.com \\
        \addr Alibaba Group, \\
        Hangzhou, China
        \AND
        \name Lu Wang \email luwang@ccs.neu.edu \\
        \addr 
        Northeastern University \\
        Boston, MA 02115}
 
\maketitle

\begin{abstract}
Automatic identification of mentioned entities in social media posts facilitates quick digestion of trending topics and popular opinions. 
Nonetheless, this remains a challenging task due to limited context and diverse name variations.
In this paper, we study the problem of entity linking for Chinese news comments given mentions' spans. We hypothesize that comments often refer to entities in the corresponding news article, as well as topics involving the entities. 
We therefore propose a novel model, \textsc{\method}, that leverages attention mechanisms to (1) pinpoint relevant context within comments, and (2) detect supporting entities from the news article.
To improve training, we make two contributions: (a) we propose a supervised attention loss in addition to the standard cross entropy, and (b) 
we develop a weakly supervised training scheme to utilize the large-scale unlabeled corpus. 
Two new datasets in entertainment and product domains are collected and annotated for experiments. 
Our proposed method outperforms previous methods on both datasets. 
\end{abstract}

\section{Introduction}
\label{sec:intro}
Social media, including online discussion forums and commenting systems, provide convenient platforms for the public to voice opinions and discuss trending events~\cite{o2010tweets,lau-collier-baldwin:2012:PAPERS}. 
Entity linking (EL), which aims to identify the knowledge base entry (or the lack thereof) for a given mention's span, has become an indispensable tool for consuming the enormous amount of social media posts.
Concretely, automatically recognizing entities can quickly inform who and what are popular~\cite{o2010tweets,zhao2014we,W16-6204}, promote semantic understanding of social media content, and facilitate downstream tasks, such as relation extraction, opinion mining, questions answering, and personalized recommendation~\cite{messenger2011recommendations, galli2015analysis}. 
Although EL has been extensively investigated in newswire~\cite{kazama-torisawa:2007:EMNLP-CoNLL20072,ratinov-EtAl:2011:ACL-HLT2011}, web pages~\cite{demartini2012zencrowd}, and broadcast news~\cite{N15-1024}, 
its study in the social media domain was only started more recently
~\cite{guo-chang-kiciman:2013:NAACL-HLT,yang-chang:2015:ACL-IJCNLP,N18-1078},
mostly focusing on English.

\begin{figure}[t]
\centering
    \includegraphics[width=152mm]{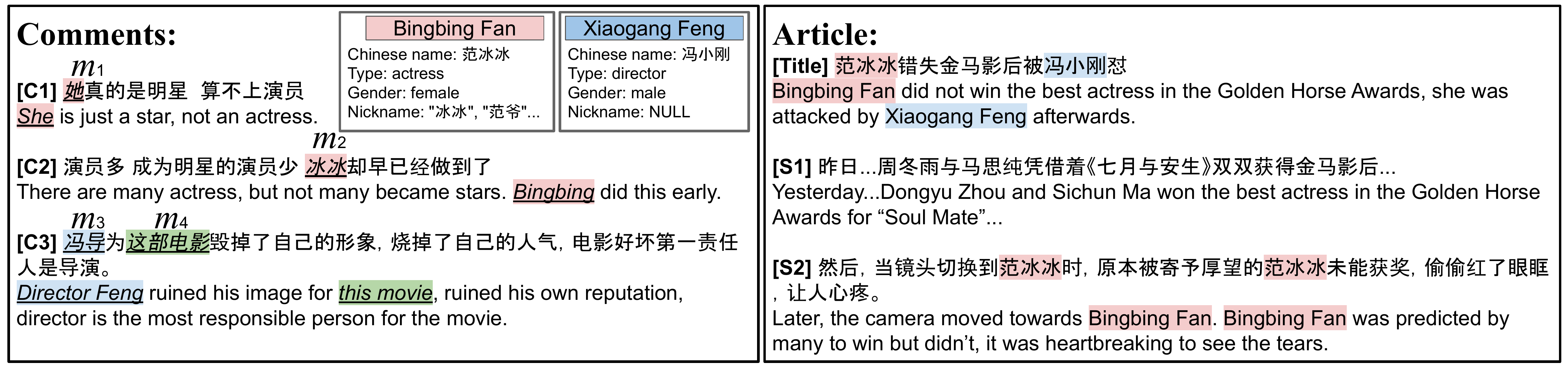}
    % \vspace{-5mm}
    \caption{\fontsize{10}{12}\selectfont 
    Sample user comments with entity mentions underlined. 
    $m_1$ (pronoun) and $m_2$ (nickname) are linked to ``Bingbing Fan''. 
    Mention $m_3$ (unknown nickname) refers to ``Xiaogang Feng''. 
    Their associated entities can be inferred from the article, but not from the comment alone. 
    }
    \label{fig:motivating_example}
    % \vspace{-5mm}
\end{figure}

In this paper, \textit{we study the task of entity linking for user comments in online Chinese news portals.} To the best of our knowledge, we are the first to investigate EL problem for the genre of news comments at a large scale. 
Besides issues present in the conventional EL work~\cite{ji2010overview}, social media text poses additional challenges: \textit{the lack of context} and \textit{increased name variations due to its informal style}. 
State-of-the-art EL methods~\cite{francislandau-durrett-klein:2016:N16-1,gupta-singh-roth:2017:EMNLP2017} heavily rely on modeling the text surrounding the mentions,
as the abundant context from longer documents greatly helps identify entity related content. 
However, context is often scant for user comments. For instance, as shown in Figure~\ref{fig:motivating_example}, the entity mention $m_2$ may indicate ``Bingbing Fan" or ``Bingbing Li", both being prominent actresses. By looking up the entities covered in the article, which contains unambiguous mention of the former entity, an EL system will be more confident to link $m_2$ to it. 
Moreover, the informal style and evolving vocabulary on social media lead to enormous name variations based on aliases, morphing, and misspelling. For instance, in one of our newly annotated datasets, the maximum number of distinct mentions of an entity is $121$.

\begin{figure*}[t]
\centering
  \includegraphics[width=150mm]{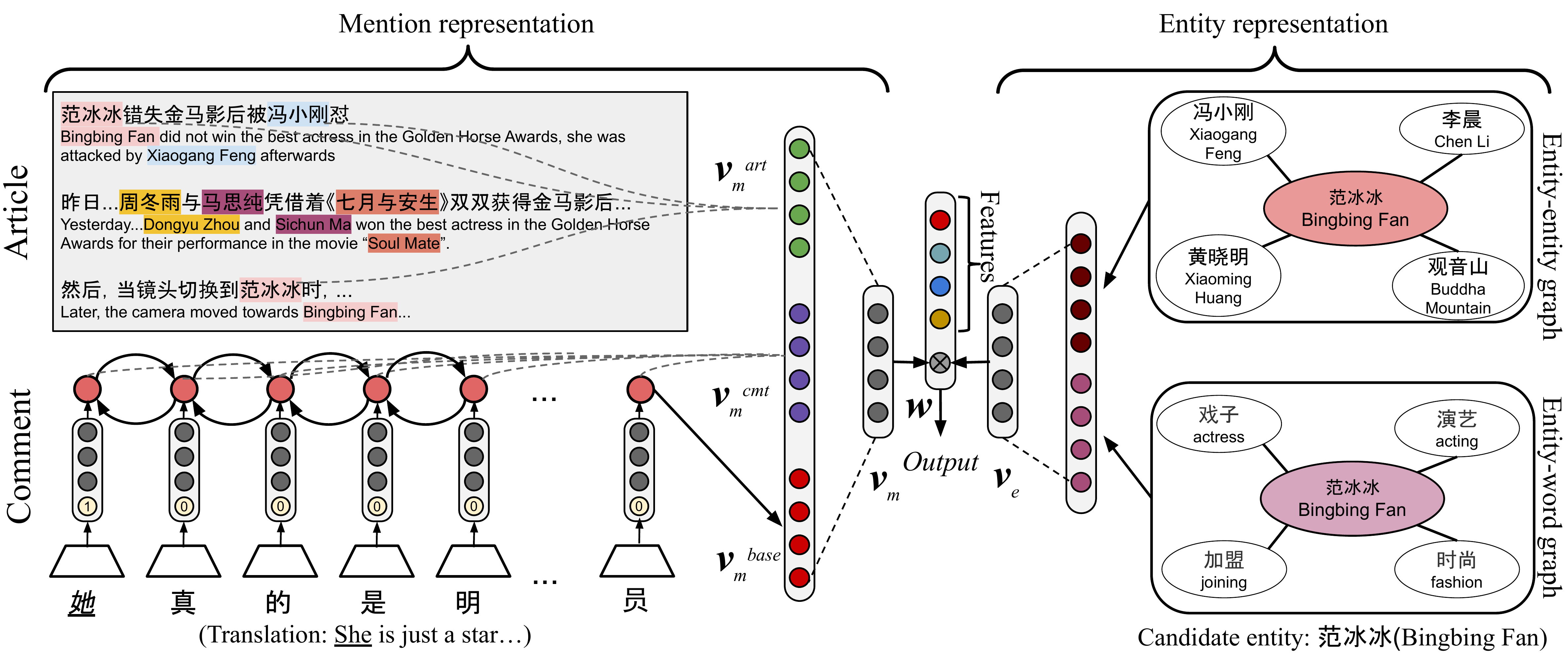}
  \caption{ 
  \fontsize{10}{12}\selectfont 
   Overview of \textsc{\method} model. It learns to represent mentions (left) and entities (right). 
   Context-aware mention representation encodes the information about comment ($v_m^{base}$ and $v_m^{cmt}$) and article ($v_m^{art}$) via attention mechanisms. Entity representation is built on entity-entity and entity-word co-occurrence graph embeddings. The dot product of the mention representation and entity representation can be concatenated with a feature vector to produce the final output after a layer of linear transformation. 
  }
   \label{fig:model-overview}
\end{figure*}

In this work, we propose \textsc{\method}, {\it a novel entity linking model for Chinese news comments by exploiting context information of entity mentions as well as identifying relevant entities in reference articles.}  
\textsc{\method}, with its overview displayed in Figure~\ref{fig:model-overview}, has three key properties. 
First, we enrich the mention representation with two sources of information through attentions. {\bf Comment attention} pinpoints topics involving the target entity from comment context. For instance, words ``star" and ``actress" in comment $C1$ in Figure~\ref{fig:motivating_example} provide useful information about entity types. {\bf Article entity attention} detects target entities from the articles if they are discussed. 
Furthermore, we investigate a new objective function to drive the learning of article entity attention.
Finally, we also exploit data augmentation with distant supervision~\cite{P09-1113} to leverage large amounts of unlabeled comments and articles for model training. 

Since there was no publicly-available annotated dataset, as part of this study, we collect and label two new datasets of Chinese news comments from the domain of entertainment and product, which are crawled from a popular Chinese news portal \url{toutiao.com}.\footnote{Datasets and code can be found at \url{http://xinyuhua.github.io/Resources/akbc20/}.} 
Experimental results show that our best performing model obtains significantly better accuracy and Mean Reciprocal Rank scores than the state-of-the-art~\cite{P18-1148} and other competitive comparisons. 
For example, our model improves the accuracy by at least three points over the state-of-the-art model in both domains with NIL mentions considered ($67.2$ vs $58.6$ on entertainment comments, and $77.3$ vs $68.6$ on product comments).

\section{Related Work}
\label{sec:related}
Entity linking (EL), as a fundamental task for information extraction, has been extensively studied for long documents, such as news articles or web pages~\cite{ji2010overview,shen2015entity}. 
State-of-the-art EL systems rely on extensive resources for learning to represent entities with diverse information, including entity descriptions given by Wikipedia or knowledge bases~\cite{kazama-torisawa:2007:EMNLP-CoNLL20072,cucerzan:2007:EMNLP-CoNLL2007}, entity types and relations with other entities~\cite{E06-1002,hoffart-EtAl:2011:EMNLP,kataria2011entity}, and the surrounding context~\cite{ratinov-EtAl:2011:ACL-HLT2011,sun2015modeling}. Neural network-based models are designed to learn a similarity measure between a given entity mention and previously acquired entity representation~\cite{francislandau-durrett-klein:2016:N16-1,gupta-singh-roth:2017:EMNLP2017,P18-1148}. However, very limited context is provided in social media posts.

In this work, we propose to leverage attention mechanisms to identify salient content from both comments and the corresponding articles to enrich the entity mention representation.

Our work is inline with the emerging entity linking research for social media content~\cite{liu-EtAl:2013:ACL20133,guo-chang-kiciman:2013:NAACL-HLT,guo-EtAl:2013:EMNLP, Q14-1021, hua2015microblog,yang-chang:2015:ACL-IJCNLP}. 
To overcome the lack of context, existing models mostly resort to including extra information, e.g., considering historical messages by the same authors or socially-connected authors~\cite{guo-EtAl:2013:EMNLP,shen2013linking,yang-chang-eisenstein:2016:EMNLP2016}, or leveraging posts of similar content~\cite{huang-EtAl:2014:P14-11}. 
However, users in news commenting systems might be anonymous, and few additional posts would be available for newly published articles. We therefore study a more practical setup, without using any of the aforementioned information as input.

\section{Data Collection and Annotation}
\label{sec:data}
\begin{table}[t]
\fontsize{10}{11}\selectfont
    \centering
    \setlength{\tabcolsep}{2mm}
    \begin{tabular}{lll}
    \toprule
    	& \textbf{Entertainment} & \textbf{Product} \\
    \midrule
    \# News Articles & 10,845 & 8,275 \\
    Avg \# Sents per Article & 18.5 & 13.5 \\
    Avg \# Chars per Sentence & 40.4 & 50.7 \\
    \midrule
    \# Comments & 967,763 & 410,790 \\
    Avg \# Chars per Comment & 21.3 & 23.2 \\
    \# Annotated Comments & 30,630 & 5,189 \\
    \# Annotated Mentions & 46,942 & 7,497 \\
    \# Annotated Unique Entities & 1,846 & 470 \\
    \bottomrule
    \end{tabular}
    \caption{\fontsize{10}{12}\selectfont 
%     Basic statistics of the dataset on two domains.
	Statistics of crawled datasets from entertainment  and product domains.
    }
    % \vspace{-5mm}
    \label{tab:basic-stats-dataset}
\end{table}

We collect user comments along with corresponding news articles from \url{toutiao.com}, a popular Chinese online news portal. 
A sample article snippet with comments is displayed in Figure~\ref{fig:motivating_example}. 
Two popular domains are selected for annotation: entertainment (\textsc{Ent}) and product (\textsc{Prod}). Articles and comments in \textsc{Ent} focus on movies, TV shows, and celebrities, whereas most topics in \textsc{Prod} are automobiles and electronic products.
The statistics of the crawled dataset after filtering are in Table~\ref{tab:basic-stats-dataset}.
As illustrated, there are only an average of $20$ characters in a comment, highlighting the lack of context.

\smallskip
\noindent \textbf{Annotation Procedure.} 
We randomly sample $995$ articles from \textsc{Ent} and $783$ articles from \textsc{Prod}, and annotate the corresponding user comments. 
Articles and comments that are not in the samples are used for model pre-training via data augmentation (\S~\ref{sec:data_augmentation}). 

Annotators are presented with both comments and corresponding articles during the annotation process. They first identify mention spans, where named, nominal, and pronominal mentions of entities are labeled. Each mention is then linked to an entity in a knowledge base, or labeled as \textsc{NIL} if no entry is found. 
Though not in our knowledge base, the word ``\begin{CJK*}{UTF8}{gbsn}小编\end{CJK*}'' (editor) is included as an entity due to its popularity. We also allow one mention to be linked to multiple entities, e.g. plural pronoun ``\begin{CJK*}{UTF8}{gbsn}他们\end{CJK*}'' (they/them). %, however, this is less frequent. 
Comments without any mention are discarded. 
$13$ professional annotators, who are native Chinese speakers with extensive NLP annotation experience, are hired, each annotating a different subset. An additional human annotator conducts the final check.

\smallskip
\noindent \textbf{Statistics.} 
Final statistics for the datasets are displayed in Table~\ref{tab:basic-stats-dataset}. 
On average, there are $4.4$ distinct mentions per entity, with a maximum number of $121$ for domain \textsc{Ent}. For \textsc{Prod}, the average mention number is $2.9$ with a maximum number of $48$.
Sample mentions are shown in Table~\ref{tab:ent-mention-count}. 

\begin{table}[t]
\fontsize{10}{11}\selectfont
    \centering
    \setlength{\tabcolsep}{.5mm}
    \begin{tabular}{|p{38mm} p{110mm}|}
    \hline
    Entity (uniq. mentions) & Sample Mentions \\
    % \multicolumn{2}{|c|}{\bf{Entertainment}} \\
     \hline\hline
        \begin{CJK*}{UTF8}{gbsn}范冰冰\end{CJK*} (121) \newline ``Bingbing Fan'' & ``\begin{CJK*}{UTF8}{gbsn}戏子\end{CJK*}(actress)'', ``\begin{CJK*}{UTF8}{gbsn}冰姐\end{CJK*} (sister Bing)'', ``\begin{CJK*}{UTF8}{gbsn}范\end{CJK*} (Fan)'', ``\begin{CJK*}{UTF8}{gbsn}国际女神\end{CJK*} (international goddess)'' \\
  \hline
        \begin{CJK*}{UTF8}{gbsn}那英\end{CJK*}  (111) \newline ``Ying Na'' & ``\begin{CJK*}{UTF8}{gbsn}戏子\end{CJK*} (actress)'', ``\begin{CJK*}{UTF8}{gbsn}满族后裔\end{CJK*} (descendant of Manchu people)'', ``\begin{CJK*}{UTF8}{gbsn}自个\end{CJK*} (herself)'', ``\begin{CJK*}{UTF8}{gbsn}演员\end{CJK*} (actress)'' \\
    
    % \multicolumn{2}{|c|}{\bf{Product}} \\
  \hline\hline
        \begin{CJK*}{UTF8}{gbsn}别克英朗\end{CJK*} (48) \newline ``Buick Excelle'' & ``\begin{CJK*}{UTF8}{gbsn}手动精英\end{CJK*} (stick shift elite)'', ``\begin{CJK*}{UTF8}{gbsn}这款车\end{CJK*} (this car)'', ``\begin{CJK*}{UTF8}{gbsn}我的车子\end{CJK*} (my car)'', ``\begin{CJK*}{UTF8}{gbsn}2016款英朗\end{CJK*} (2016 Excelle)'' \\
  \hline
    \begin{CJK*}{UTF8}{gbsn}马自达3昂克赛拉\end{CJK*} (47) \newline ``Mazda3 Axela'' & ``\begin{CJK*}{UTF8}{gbsn}两厢\end{CJK*} (hatchback)'', ``\begin{CJK*}{UTF8}{gbsn}自动舒适型\end{CJK*} (automatic and comfortable)'', ``\begin{CJK*}{UTF8}{gbsn}昂克塞拉\end{CJK*} (Axela)'', ``\begin{CJK*}{UTF8}{gbsn}昂克赛拉1.5自动舒适车\end{CJK*} (Axela 1.5T automatic)'' \\
    \hline
    \end{tabular}
    % \vspace{-2mm}
    \caption{\fontsize{10}{12}\selectfont 
    Entities with the most unique mentions from entertainment and product domains. 
    }
    % \vspace{-7mm}
    \label{tab:ent-mention-count}
\end{table}

\begin{table}
% \vspace{-2mm}
\centering
\fontsize{10}{11}\selectfont
\begin{tabular}{lllllll}
\toprule

& \textbf{Canon.} & \textbf{Nick.} & \textbf{Pron.} & \textbf{Others} & \textbf{Plural} & \textbf{NIL} \\
\midrule
\textbf{Entertainment} & 29.8\% & 4.0\% & 12.9\% & 21.9\% & 2.9\% & 28.4\% \\
\textbf{Product} & 33.9\% & 0.6\% & 2.7\% & 41.1\% & 0.2\% & 21.6\% \\

\bottomrule
\end{tabular}
% \vspace{-3mm}
\caption{\fontsize{10}{12}\selectfont
Mention type distribution.}
\label{tab:mention-types}
\end{table}

We categorize the samples into the following types, based on the entities mentioned by: 
(1) \textbf{canonical names} as defined in knowledge base; 
(2) \textbf{nicknames} as the popular aliases included in the knowledge base for each entity; 
(3) \textbf{pronominal} mentions indicating one entity, such as ``\begin{CJK*}{UTF8}{gbsn}他\end{CJK*} (he/him)'' or ``\begin{CJK*}{UTF8}{gbsn}这个\end{CJK*} (this)''; 
(4) \textbf{plural pronominal} mentions that are linked to multiple entities; 
(5) \textbf{others}, all other types of mentions that can be linked to the KB, including aliases not in the knowledge base or misspellings; 
and (6) \textbf{NIL}, mentions that cannot be linked to any entity in the KB. 
The mention type distributions are in Table~\ref{tab:mention-types}, and it is observed that pronominal and nickname mentions are more common in \textsc{Ent} where celebrities are frequently discussed. %in the comments. 
The type of \textit{others} is more significant in \textsc{Prod} due to the prevalent usage of irregular name variations for products.

\smallskip
\noindent \textbf{Knowledge Base.} 
Baidu Baike\footnote{\url{https://baike.baidu.com}}, 
a large-scale Chinese online encyclopedia, is used to construct the knowledge base (KB). 
A snapshot of Baike containing 68,067 unique entities was collected on May 10th, 2017.
Four attributes are leveraged for feature engineering: 
%For developing comparison methods and our models, we leverage four important attributes: 
(1) \textit{gender}, 
(2) \textit{nicknames} as a list of common aliases for an entity,
(3) \textit{entity type},
and (4) \textit{entity relation}. 

\section{The Proposed Approach}
\label{sec:model}
Our model takes as input a phrasal mention $m$ in a comment $C$, which is posted under an article. 
Given a knowledge base, we aim to predict the KB entity that $m$ refers to, or to label it as \textsc{NIL} if no such entity exists. 
Concretely, a list of candidate entities will be first selected as $\mathcal{E}_m = \{e\}$ based on string matching and knowledge graph expansion (see \S~\ref{sec:candidate}). Then a linking probability will be computed over each candidate given $m$.% the mention. 

\subsection{Candidate Construction}
\label{sec:candidate}

Our candidate construction algorithm consists of two steps. For each mention, we consider all entities that appear in the same comment and corresponding article by matching their canonical names. 
This forms the initial candidate list. 
In the second step, a new entity is selected if it has a relation with any entity in the initial list according to our KB. The initial list, the expanded entities, and NIL comprise the final candidate set. 

Following this procedure, $96\%$ gold-standard entities are retrieved in the candidate sets for \textsc{Ent}, and $62\%$ are covered for \textsc{Prod}. 
To improve the coverage for the \textsc{Prod} domain, we collect unambiguous aliases (no other entity with the same alias) that are not pronominal mentions from training data for each entity, and use these as additional entity nicknames for candidate construction. The coverage is increased to $93\%$.

\subsection{Entity Representation}
\label{sec:entity_representation}
Prior work for entity representation learning usually relies on entity-word co-occurrence statistics derived from the entities' English Wikipedia pages~\cite{francislandau-durrett-klein:2016:N16-1,gupta-singh-roth:2017:EMNLP2017,D17-1277,eshel-EtAl:2017:CoNLL}. 
Unfortunately, Wikipedia has low coverage of entities in our newly collected Chinese datasets. We thus consider two sources of information, both acquired from news headlines. First, a graph-based node2vec \cite{grover2016node2vec} embedding $\bm{u}^{nod}$ is induced from an \textit{entity-entity} co-occurrence matrix extracted from 65 million news titles after applying canonical name matching~\cite{zwicklbauer2016robust,K16-1025}. 
%note: similar work that uses entity-entity co-occurrence \cite{zwicklbauer2016robust,K16-1025}
$\bm{u}^{nod}$ is expected to capture entity relations. %, e.g., entities that are co-stars of movies, or products of similar price range.  
Second, a Singular Vector Decomposition (SVD)-based representation $\bm{u}^{wrd}$ is obtained from an \textit{entity-word} co-occurrence matrix constructed from the same set of news titles. 
We concatenate them as $\bm{u}$  and apply a one-layer feedforward neural network over it to form the entity representation $\bm{v}_e = \tanh(\bm{W}_e\bm{u} + \bm{b}_e)$,
where $\bm{W}_e \in \mathbb{R}^{300\times 600}$ and $\bm{b}_e\in \mathbb{R}^{300 \times 1}$ are trainable parameters.

\subsection{Mention Representation}
%In the base model, the mention context is represented solely by its comment.
We train character embeddings %$\bm{x}^c$ 
from the 327 million user comments with word2vec~\cite{mikolov2013efficient}. 
A bidirectional Long Short-Term Memory (biLSTM) network is then applied over comment character embeddings $\bm{x}_i^c$, with hidden state $\bm{h}_i=[\overrightarrow{\bm{h}_i}; \overleftarrow{\bm{h}_i}]$ for each time step $i$. 
%In order to distinguish different mentions in the same comment, 
We append a one-bit mask $q_i$ to the character embeddings to indicate the mention span. If a character is within the mention span, $q_i$ is $1$; otherwise, it is $0$. $\bm{h}_i$ is calculated recurrently as $\bm{h}_{i} = g(\bm{h}_{i-1}, [\bm{x}_i^c;q_{i}])$, where $g$ is the 200-dimensional biLSTM network.
The last hidden state $\bm{h}_T$ is taken as the base form of mention representation $\bm{v}_m^{base}$.

\noindent \textbf{Comment Attention.} 
Preliminary studies show that $\bm{v}_m^{base}$ focuses on the local context, and does not capture long-distance information well. %, which can be also relevant for entity linking. 
Hence we propose to learn an importance distribution over all comment characters through a bilinear attention \cite{luong-pham-manning:2015:EMNLP} with query $\bm{\tilde{m}}$, the average character embeddings of the mention: 

{
\setlength{\abovedisplayskip}{2pt}
\setlength{\belowdisplayskip}{2pt}
\begin{align}
& \bm{\tilde{m}} = \frac{1}{m_e - m_s}\sum_{i=m_s}^{m_e}{\bm{x}^c_i} \label{eq:cattn_1} \\
  & \alpha_i^{cmt} = \frac{\textnormal{exp}(\bm{h}_i^T \bm{W}_c  \bm{\tilde{m}})}{\sum_{i^\prime=1}^T\textnormal{exp}(\bm{h}_{i^\prime}^T \bm{W}_c \bm{\tilde{m}})} \\
  & \bm{v}_m^{cmt} = \sum_{i=1}^T{\alpha_i^{cmt}\bm{h}_i} \label{eq:cattn_4}
\end{align}
}
\noindent where $m_s$, $m_e$ are start and end offsets of the mention span. $\bm{x}^c_i$ is the character embedding of the $i$-th character in the comment, $\bm{W}_c \in \mathbb{R}^{200\times 300}$ is the trainable bilinear matrix.

\smallskip
\noindent \textbf{Article Entity Attention.} 
Intuitively, users tend to comment on entities covered in the news. We thus design an article entity attention to identify target entities if they appear in the article, or indicate non-existence otherwise. 
Concretely, articles are segmented into words by Jieba\footnote{\url{https://github.com/fxsjy/jieba}}, an open source Chinese word segmentation tool. 
Each word is matched with canonical entity names in the knowledge base, and the article is represented as a set of unambiguous entities, $\mathcal{E}_a$. 
Each entity is represented as $\bm{u}=[\mathbf{u}^{nod};\mathbf{u}^{wrd}]$.
We also add one {\it absent padding entity} (denoted as {\bf ABS}), a $300$-dimension zero vector, into the set to indicate that the entity is not in the article. 
The article entity representation $\bm{v}_m^{art}$ is calculated as:

{
\setlength{\abovedisplayskip}{2pt}
\setlength{\belowdisplayskip}{2pt}
\begin{align}
 & \beta_{j}^{art} = \text{softmax}(\bm{u}_j^T\bm{W}_a \bm{\tilde{m}}) \label{eq:aattn_3} \\
  & \bm{v}_m^{art} = \sum_{j=1}^{|\mathcal{E}_a|}{\beta_{j}^{art}\bm{u}_j} \label{eq:aattn_4}
\end{align}
}
where $\bm{u}_j$ is the entity representation for $j$-th entity in $\mathcal{E}_a$. 
$\bm{W}_a \in \mathbb{R}^{600 \times 300}$ is the bilinear matrix parameter.

\subsection{Learning Objective}
\textsc{\method} learns to align the mention representation $\bm{v}_m$ and the candidate entity representation $\bm{v}_e$ after transforming them into a common semantic space. 
Specifically, the base form $\bm{v}_{m}^{base}$, comment attended $\bm{v}_m^{cmt}$, and article attended $\bm{v}_{m}^{art}$ are concatenated as the input to a feedforward neural network to form $\bm{v}_{m}$ as $\bm{v}_m =\tanh(\bm{W}_m[\bm{v}_{m}^{base}; \bm{v}_{m}^{cmt}; \bm{v}_{m}^{art} \big ] + \bm{b}_m)$.
Given a mention $m$ represented as $\bm{v}_{m}$, the probability for $m$ being linked to an entity $e$ (represented as $\bm{v}_e$) is computed by applying the softmax function over the dot product between their representations, over all candidates in $\mathcal{E}_m$: $P(e|m) = \text{softmax}_{e \in \mathcal{E}_m}(\bm{v}_e \cdot \bm{v}_m)$
The entity with the highest positive likelihood is selected as prediction. 
Previous work~\cite{yang-chang-eisenstein:2016:EMNLP2016} has found that surface features can further improve representation learning-based EL models. We thus append {\bf features} (\S~\ref{subsec:features}) to the dot product via {\small $P(e|m) = \text{softmax}_{e \in \mathcal{E}_m}\Big ( \mathbf{w}\cdot \big [\bm{v}_e \cdot \bm{v}_m ; \mathbf{\Phi}(m)\big ] \Big )$}, where $\mathbf{\Phi}(m)$ is the feature vector and $\mathbf{w}$ are learnable weights.

During training time, we use the same candidate construction algorithm in \S~\ref{sec:candidate} to collect negative samples, where all candidates except the gold-standard are treated as negative. 
The cross-entropy loss on training set is defined as: 

% \vspace{-2mm}
{
% \fontsize{9}{11}\selectfont
\setlength{\abovedisplayskip}{2pt}
\setlength{\belowdisplayskip}{2pt}
\begin{align}
&\mathcal{L}_{EL}(\theta) = -\sum_n\sum_{k} y_{n,k}^*\log(P(e_k | m_n))
\end{align}
}
\noindent 
where $P(e_k|m_n)$ is the predicted probability for the $k$-th entity candidate for $n$-th mention in training set. $y_{n,k}^*$ represents the gold-standard, it has a value of $1.0$ for positive samples, and $0.0$ for negative ones.

\paragraph{Supervised Attention Loss.}
Notice that the article entity attention naturally learns an alignment between the mention and entity representation $\bm{u}$. To help learn high quality alignment, we design a new learning objective to provide direct supervision to the article entity attention. To the best of our knowledge, we are the first to design supervised attention mechanism to guide entity linking. 
Concretely, during training, if an entity in $\mathcal{E}_a$ matches the gold-standard, we assign a relevance value of $1.0$ to it; otherwise, the score is $0.0$. If none from $\mathcal{E}_a$ matches, the absent padding entity is labeled as relevant. 
We thus design the following objective for article attention learning:

{
\setlength{\abovedisplayskip}{2pt}
\setlength{\belowdisplayskip}{2pt}
\begin{align}
\begin{split}
\mathcal{L}_{Att}(\theta)  = & -\sum_n\sum_{j} \beta_{n,j}^*\log(\hat{\beta}_{n,j})  
\label{eq:loss_entity} \\
\end{split}
\end{align}
}
\noindent 
$\beta^*_{n,j}$ is the true relevance value for $j$-th article entity, and $\hat{\beta}_{n,j}$ is the attention calculated as in Eq.~\ref{eq:aattn_3}, both are extended with mention index $n$ (i.e. the $n$-th mention in the training set).
The {\bf final learning objective} becomes $\mathcal{L}(\theta)  = \mathcal{L}_{EL}(\theta) + \lambda \cdot \mathcal{L}_{Att}(\theta)$. $\lambda$ is set to $0.1$ in all experiments below. 

\subsection{Features}
\label{subsec:features}
We optionally append $20$ features to the output layer, as detailed in Table~\ref{tab:features}, where the last $11$ features are adopted from ~\cite{N10-1072}.
\begin{table}[h]
  \fontsize{9}{11}\selectfont
    \centering
    \begin{tabular}{p{21mm}p{120mm}}
 \toprule
    \rowcolor{Gray}
        {\bf Feature} & {\bf Description} \\
        CanonMatch & Whether the mention text exact-match the canonical KB name \\
        \rowcolor{Gray}
        NicknMatch & Whether the mention text exact-match the canonical KB name \\
        CharJaccard & The char-level Jaccard score btw. the mention and entity's canonical name \\
        \rowcolor{Gray}
        PinyJaccard & The Jaccard similarity between the Pinyin of the mention and entity's canonical name \\
        GendMatch & Whether the gender of pronominal mention matches that in KB \\
        \rowcolor{Gray}
        EntArtFreq & The frequency of candidate entity in article, considering both exact canonical name searching and nickname searching \\
        CommentDist & The distance between the canonical name of the candidate entity and the mention in the comment, if the canonical name is not present set to 100 \\
        \rowcolor{Gray}
        PriorProb & probability $P(e|m)$ with MLE \\
        Special & Whether the mention is a domain-specific entity, such as ``\begin{CJK*}{UTF8}{gbsn}小编\end{CJK*}'' (editor) \\
        \rowcolor{Gray}
        EditDist & The edit distance between mention and entity on character level \\
        StartWithMent & Whether any of the entity's canonical name or nickname starts with the mention string \\
        \rowcolor{Gray}
        EndWithMent & Whether any of the entity's canonical name or nickname ends with the mention string \\
        StartInMent & Whether any of the entity's canonical name or nickname is a prefix of the mention string\\
        \rowcolor{Gray}
        EndInMent & Whether any of the entity's canonical name or nickname is an affix of the mention string\\
        EqualWordCnt & The maximum number of same words between mention and entity's canonical name and nicknames \\
        \rowcolor{Gray}
        MissWordCnt & The minimum number of different words between mention and entity's canonical name and nicknames \\
        ContxtSim & TF-IDF similarity between entity's Baike article and comment \\
        \rowcolor{Gray}
        ContxtSimRank & Inverted rank of ContxtSim across all candidates\\
        AllInSrc & Whether  all words in candidate entity's canonical name exist in comment \\
        \rowcolor{Gray}
        MatchedNE & The number of matched named entities between entity's Baike page and comment \\
    \bottomrule
    \end{tabular}
    % \vspace{-3mm}
     \caption{\fontsize{10}{12}\selectfont 
    Features used in our model and comparisons.
    }
    % \vspace{-5mm}
    \label{tab:features}
\end{table}

\subsection{Weakly Supervised Pre-training} 
\label{sec:data_augmentation}
We leverage the unlabeled samples for data augmentation. Concretely, mentions and entities are automatically labeled if an entity's canonical name or nickname can be matched in a comment unambiguously (i.e., no other entity with the same name). 
In total, this procedure automatically labeled $502,858$ comments for the \textsc{Ent} domain, which is split into $453,080$ for training and $49,778$ for validation. For the \textsc{Prod} domain, we create $175,951$ comments, among which $158,336$ are for training and $17,615$ are for validation. Each dataset is used to pre-train \textsc{\method}, which is then trained on the annotated data.

\section{Experimental Setup}
\label{sec:experiment}
Each dataset is split into training, validation, and test sets based on articles, with statistics displayed in Table~\ref{tab:split-data}.
Articles in test sets are published later than those in training and validation sets. 
For this study, we focus on the task of entity linking, therefore gold-standard mention spans are assumed to have been provided. A mention detection component will be developed in future work. 

\begin{table}
\fontsize{10}{11}\selectfont
    \centering
    \setlength{\tabcolsep}{1.2mm}
    \begin{tabular}{lllcllcll}
    \toprule
    	& \multicolumn{2}{c}{\bf{Train}} & \phantom{} & \multicolumn{2}{c}{\bf{Valid}} & \phantom{} & \multicolumn{2}{c}{\bf{Test}} \\
    % 	\cmidrule{2}\cmidrule{4}\cmidrule{6}
        & article &  comment & & article & comment && article & comment \\
         \midrule
        \textbf{Entertainment} & 734 & 23,046 && 98 & 3,153 && 149& 4,415 \\
        \textbf{Product} & 587 & 3,943 && 78 & 473 && 118& 773 \\
         \bottomrule
    \end{tabular}
    % \vspace{-3mm}
    \caption{\fontsize{10}{12}\selectfont 
    Experimental setup statistics. 
    }
    % \vspace{-4mm}
    \label{tab:split-data}
\end{table}
\smallskip
\noindent \textbf{Hyperparameters.}
For all experiments, Adam optimizer \cite{kingma2014adam} is used with an initial learning rate of $0.0001$. We adopt gradient clipping with a maximum norm of $5$. Model batch size is set to $128$. 

\smallskip
\noindent \textbf{Baselines.} 
We design five baselines: 
(1) \textsc{MatchCanon} matches the mention with canonical names in KB, and outputs an entity if a match is found, otherwise predicts NIL; 
(2) \textsc{MatchCanonAndNick} further matches nicknames if \textsc{MatchCanon} returns NIL, 
(3) \textsc{FrequencyInArt} predicts the most frequent entity in the article; 
(4) \textsc{FirstInArt} predicts the first entity in the article;
(5) \textsc{PriorProb} predicts the most likely entity based on $P(e|m)$, estimated from entity-mention co-occurrence in the training set.

\smallskip
\noindent \textbf{Comparisons.} 
We further compare against the following models: 
(1) Vector Space Model (\textsc{VSM}) %~\cite{bagga1998entity} 
computes TF-IDF cosine similarity between mention context and entity KB pages~\footnote{We include all content from entity's Baidu Baike page.}, with the most similar candidate as prediction. 
(2) Logistic Regression (\textsc{LogReg}) trained with features described in the next paragraph.
(3) \textsc{ListNet} is a learning-to-rank approach that outperforms all methods in the EL track of TAC-KBP2009~\cite{N10-1072}. 
(4) \textsc{CEMEL} expands mention representation with similar posts and then applies \textsc{VSM} ~\cite{guo-EtAl:2013:EMNLP}. We retrieve all comments containing the mention string from the training set as similar posts. 
(5) \textsc{Ment-norm}~\cite{P18-1148} is the state-of-the-art EL model on AIDA-CoNLL~\cite{D11-1072}, which consists of English news articles. 
It leverages latent relations among mentions to find global optimal linking results. Important parameters of \textsc{Ment-norm}, such as the number of latent relations, are tuned on our development set. The same entity and character embeddings as in our model are utilized.

\section{Results and Analysis}
\label{sec:analysis}

\begin{table*}[t]
\fontsize{9}{10}\selectfont
    \centering
    \setlength{\tabcolsep}{.7mm}
    \begin{tabular}{lllllcllll}
    \toprule
    & \multicolumn{4}{c}{\textbf{Entertainment}} & \phantom{} & \multicolumn{4}{c}{\textbf{Product}} \\
    \cmidrule{2-5}\cmidrule{7-10}
    	& \multicolumn{2}{c}{\textit{with NIL mentions}} & \multicolumn{2}{c}{\textit{w/o NIL mentions}} &
    	\phantom{} & \multicolumn{2}{c}{\textit{with NIL mentions}} & \multicolumn{2}{c}{\textit{w/o NIL mentions}}  \\
        & \textbf{Acc} & \textbf{MRR} & \textbf{Acc} &\textbf{MRR} & \phantom{} & \textbf{Acc} & \textbf{MRR} & \textbf{Acc} &\textbf{MRR} \\
        \midrule
        \multicolumn{9}{l}{\bf Baselines}\\
        \textsc{MatchCanon}  & 42.70  & -& 37.91  & - & \phantom{} &  54.35  & - & 44.00  & - \\
        \textsc{MatchCanonAndNick}  & 44.66 & -  & 40.22 & - & \phantom{} & 54.97  & - & 44.78  & -  \\
        \textsc{FrequencyInArt}  & 32.66 & -  & 35.67 & - &\phantom{} &  16.34 &  - & 20.44 &  -   \\
        \textsc{FirstInArt}  & 29.17 & -  & 31.86 & - &\phantom{} &  17.41 & -  & 21.78 & -\\
        \textsc{PriorProb}  & 54.22 & 57.16  & 51.65 & 54.85 &\phantom{} &  77.71 & 78.44  & 76.89 & 77.80  \\
        \midrule
        \multicolumn{9}{l}{\bf Learning-based Models}\\
        \textsc{VSM}  & 26.04 & 34.70  & 28.44 & 37.90  &\phantom{} &  33.57 & 42.72& 42.00 & 53.45  \\
        \textsc{LogReg} & 57.69 & 62.76  & 63.00 & 68.54  &\phantom{} &  61.37 & 63.11 & 76.78 & 78.96   \\
        \textsc{Listnet}  & 55.44 & 59.22  & 60.55 & 64.67 &\phantom{} &  61.55 & 63.18 & 77.00 & 79.05 \\
        \textsc{CEMEL}   & 33.01 & 39.48 & 36.05 & 43.11  & \phantom{} &  44.23 & 50.31  & 55.33 & 62.94 \\
        \textsc{Ment-norm}   & 58.60 & 63.69& 60.53 & 65.50 & \phantom{} & 68.56 & 71.32  & 79.22 & 81.81   \\
        {\bf \textsc{XREF}} (Ours) & {\bf 67.22}* & {\bf 73.92}* & {\bf 69.84*} & {\bf 75.46*}  &\phantom{} &  {\bf 77.26} & {\bf 81.52}  & {\bf 81.56} & {\bf 83.94} \\
        
        \bottomrule
    \end{tabular}
    % \vspace{-3mm}
    \caption{\fontsize{10}{12}\selectfont 
    Entity linking results on singular mentions with and without NIL (non-existence in KB) considered. 
    The best performing learning-based models are highlighted in \textbf{bold} per column. No MRR result is reported for baselines where only one entity is returned. Our models that are statistically significantly better than all the baselines and comparisons are marked with $\ast$ ($p < 0.0001$, approximation randomization test~\cite{noreen1989computer}).
    }
    % \vspace{-7mm}
    \label{tab:main-results}
\end{table*}

\noindent \textbf{Main Results.}
We report evaluation results based on accuracy and Mean Reciprocal Rank (MRR)~\cite{voorhees1999trec}, which considers the positions of gold-standard entities ranked by each system. 
Table~\ref{tab:main-results} displays evaluation results for entity mentions excluding plural pronominal mentions. We experiment with two setups based on whether NIL %(non-existence of entities in KB) 
is considered for training and prediction. 

Overall, our model achieves significantly better results than all other comparisons on the \textsc{Ent} domain for both setups ($p < 0.0001$, approximation randomization test). 
For \textsc{Prod} domain, our model also obtains the best accuracy and MRR when NIL is not included. 
When NIL is considered, while the strong baseline based on prior probability $p(e|m)$ achieves marginally better accuracy, our model yields higher MRR. 
This is because the \textsc{Ent} domain has much more pronominal mentions ($12.9\%$) than the \textsc{Prod} domain ($2.7\%$). On \textsc{Ent}, our models perform especially well at resolving pronominal cases; on \textsc{Prod}, the prior baseline memorizes the names better, yet our model still obtains the best MRR when NIL is considered.

\smallskip
\noindent \textbf{Results on Plural Pronominal Mentions.} 
Though rarely studied in prior work~\cite{ji2016overview}, it is common to observe pronominal mentions linked to multiple entities in social media. Here we report results on plural pronominal mentions only in Table~\ref{tab:plural-results}. We assume the true number of entities is given as $K$, which varies among samples; top $K$ candidates output by each model are compared against the gold-standards. In addition to accuracy@$K$ and MRR, Normalized Discounted Cumulative Gain (NDCG) \cite{jarvelin2002cumulated} that considers multiple target predictions is reported. 
As can be seen, \textsc{\method} with article entity attention significantly outperforms other comparisons. This is likely because plural nominal often refers to the entities in the article, suggesting the effectiveness or article attentions in these samples. Further experiments show that the full model with additional comment attention and features actually yields marginally lower scores.

\begin{table}
\fontsize{10}{11}\selectfont
    \centering
     \setlength{\tabcolsep}{1.2mm}
    \begin{tabular}{llll}
    \toprule
    	& \textbf{Acc@$K$} & \textbf{MRR} & \textbf{NDCG} \\
    	\midrule
    	\textsc{Listnet} & 1.97 & 31.70 & 42.59 \\
    	\textsc{Ment-norm} & 4.72 & 33.98  & 34.74 \\
        \textsc{LogReg} & 12.99 & 55.07 & 60.69 \\
        {\bf \textsc{XREF}} w/ Art Attn & {\bf 30.31}$^\ast$ & {\bf 63.76}$^\ast$ & {\bf 68.61}$^\ast$\\
        \bottomrule
        \end{tabular}
    % \vspace{-3mm}
    \caption{\fontsize{10}{12}\selectfont 
    Results on plural pronominal mentions for \textsc{Ent} domain. 
    $K$ indicates the number of entities in the gold-standard. 
    Significant better results than all comparisons are marked with $\ast$ ($p < 0.0001$, approximation randomization test).
    }
    \label{tab:plural-results}
\end{table}

\begin{figure}[t]
    \centering
    % \vspace{-2mm}
    \includegraphics[width=100mm]{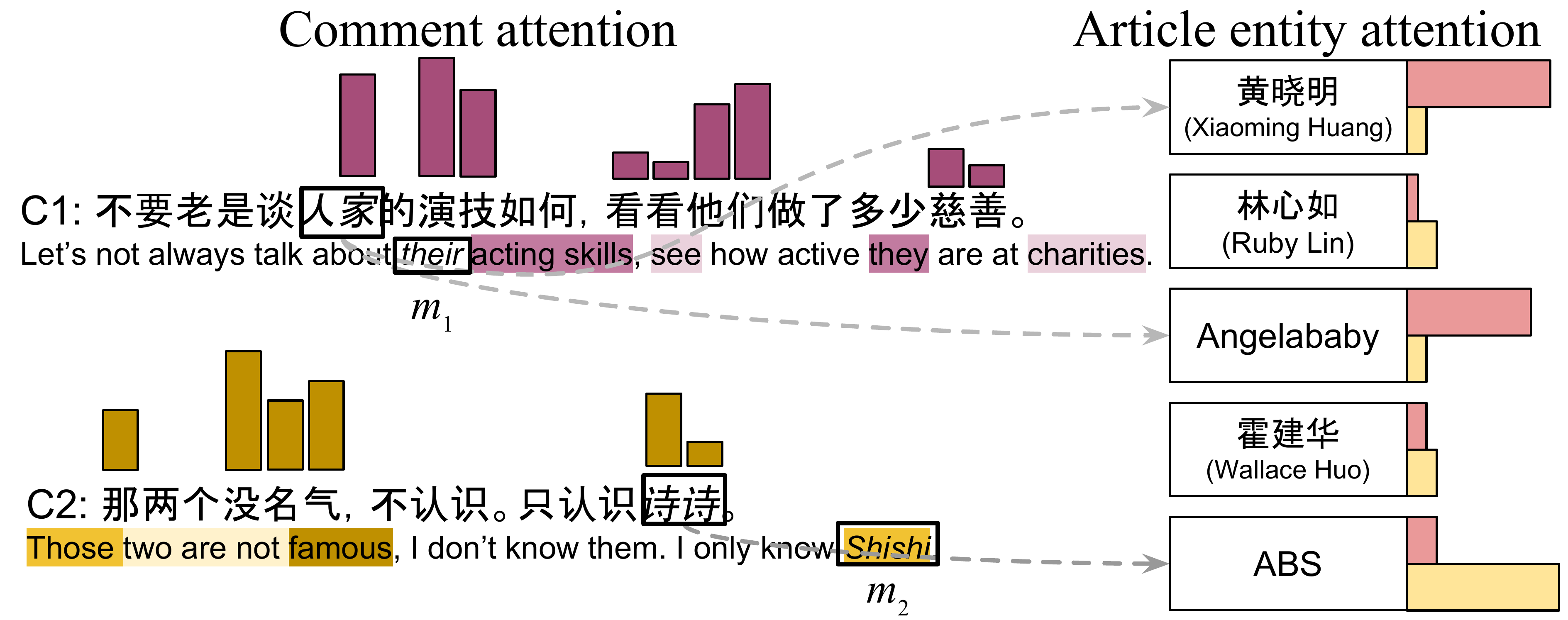}
    \caption{
    \fontsize{10}{12}\selectfont 
    An illustration of comment and article attentions. Attention over comment characters is depicted by color shading. Note that the article attention successfully selects both entities for $m_1$ (``their") (pink histograms). ABS: absent padding entity, indicating entity not in article.}
    % \vspace{-3mm}
    \label{fig:sample-attetion}
\end{figure}

\begin{figure}[t]
    \centering
    
    \includegraphics[width=150mm]{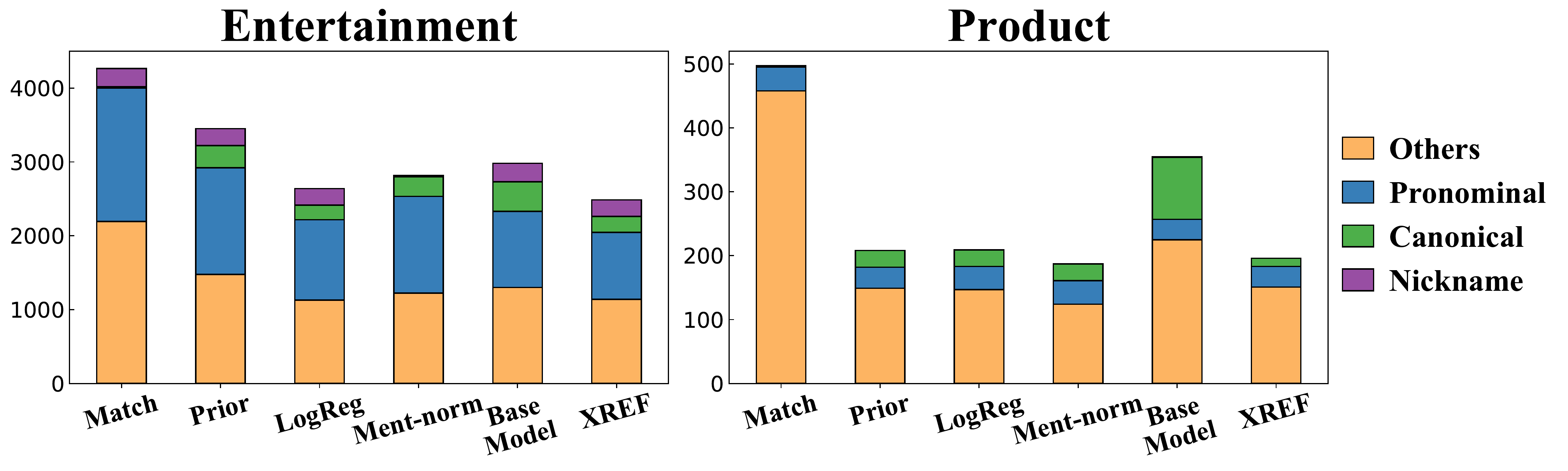}
 
    \caption{
    \fontsize{10}{12}\selectfont 
    Error breakdown based on mention type. Our model makes less errors on pronominal mentions and name variations (Others) not captured by KB. ``Base Model" represents \textsc{Xref} without attentions and features.}
    \label{fig:error-analysis}
\end{figure}

We further show sample comment attention and articles entity attention output by our model in Figure~\ref{fig:sample-attetion}. 
For the plural pronominal mention (``their") in comment $C1$, the article entity attention correctly identifies both ``Xiaoming Huang" and ``Angelababy" from the news. Comment attention also pinpoints phrases related to the entities, e.g., ``acting skills" and ``charities". For mention $m_2$, the article entity attention also correctly indicates entity's non-existence in the article by giving a high weight to the absent padding entity.

\smallskip
\noindent \textbf{Error Analysis.} 
We break down the errors made by each model based on different \emph{mention types}, as illustrated in Figure~\ref{fig:error-analysis}. 
Our model \textsc{\method} produces much less errors in pronominal mentions and other name variations than the comparisons. 
However, the name matching-based baseline achieves better performance on canonical mentions, indicating a future direction for designing better representation learning over names. 

\smallskip
\noindent \textbf{Effect of Data Augmentation and Ablation Study.} 
We examine the effect of data augmentation by evaluating models that are trained with manually labeled data only. As can be seen in Table \ref{tab:data-aug-comparison}, for both domains, there are significant accuracy drops. %---a decrease of $7.2$ points for \textsc{Ent} and $12.9$ for \textsc{Prod}. 
Moreover, accuracy drops further when attentions or features are removed. %: $15.1$ for \textsc{Ent} and $23.0$ for \textsc{Prod}.
This again demonstrates the effectiveness of comment attention and article entity attention proposed by this work.

\begin{table}[t]
\fontsize{10}{11}\selectfont
    %\hspace{-2mm}
    \centering
    \setlength{\tabcolsep}{1.2mm}
    \begin{tabular}{lll}
    \toprule
    	& \multicolumn{1}{c}{\textbf{Entertainment}} & \multicolumn{1}{c}{\textbf{Product}} \\
    	\midrule
        % {\bf \textsc{\method}} & 57.93  & 65.33  \\
        {\bf \textsc{\method}} & 57.93  & 65.33  \\
        \quad w/o Comment Attn & 51.94  & 62.22 \\
        \quad w/o Comment $+$ Article Attn & 44.90  & 55.33 \\
        
        \bottomrule
    \end{tabular}
    % \vspace{-3mm}
    \caption{\fontsize{10}{12}\selectfont 
    Accuracy by our models without data augmentation (NIL not considered). 
    }
    % \vspace{-3mm}
    \label{tab:data-aug-comparison}
\end{table}

%\vspace{-1mm}
\section{Conclusion}
\vspace{-1mm}
\label{sec:conclusion}
We present a novel entity linking model, \textsc{\method}, for Chinese online news comments. Attention mechanisms are proposed to identify salient information from comments and corresponding article to facilitate entity resolution. Model pre-training based on data augmentation is conducted to improve performance. 
Two large-scale datasets are annotated for experiments. 
Results show that our model significantly outperforms competitive comparisons, including previous state-of-the-art. 
For future work, additional languages, including low-resource ones, will be investigated.

\bibliography{ref}

\begin{thebibliography}{45}
\providecommand{\natexlab}[1]{#1}
\providecommand{\url}[1]{\texttt{#1}}
\expandafter\ifx\csname urlstyle\endcsname\relax
  \providecommand{\doi}[1]{doi: #1}\else
  \providecommand{\doi}{doi: \begingroup \urlstyle{rm}\Url}\fi

\bibitem[Benton and Dredze(2015)]{N15-1024}
Adrian Benton and Mark Dredze.
\newblock Entity linking for spoken language.
\newblock In \emph{Proceedings of the 2015 Conference of the North American
  Chapter of the Association for Computational Linguistics: Human Language
  Technologies}, pages 225--230. Association for Computational Linguistics,
  2015.
\newblock \doi{10.3115/v1/N15-1024}.
\newblock URL \url{http://aclweb.org/anthology/N15-1024}.

\bibitem[Bunescu and Pa{\c{s}}ca(2006)]{E06-1002}
Razvan Bunescu and Marius Pa{\c{s}}ca.
\newblock Using encyclopedic knowledge for named entity disambiguation.
\newblock In \emph{11th Conference of the European Chapter of the Association
  for Computational Linguistics}, 2006.
\newblock URL \url{http://aclweb.org/anthology/E06-1002}.

\bibitem[Cucerzan(2007)]{cucerzan:2007:EMNLP-CoNLL2007}
Silviu Cucerzan.
\newblock Large-scale named entity disambiguation based on {Wikipedia} data.
\newblock In \emph{Proceedings of the 2007 Joint Conference on Empirical
  Methods in Natural Language Processing and Computational Natural Language
  Learning (EMNLP-CoNLL)}, pages 708--716, Prague, Czech Republic, June 2007.
  Association for Computational Linguistics.
\newblock URL \url{http://www.aclweb.org/anthology/D/D07/D07-1074}.

\bibitem[Demartini et~al.(2012)Demartini, Difallah, and
  Cudr{\'e}-Mauroux]{demartini2012zencrowd}
Gianluca Demartini, Djellel~Eddine Difallah, and Philippe Cudr{\'e}-Mauroux.
\newblock Zencrowd: leveraging probabilistic reasoning and crowdsourcing
  techniques for large-scale entity linking.
\newblock In \emph{Proceedings of the 21st international conference on World
  Wide Web}, pages 469--478. ACM, 2012.

\bibitem[Dredze et~al.(2016)Dredze, Andrews, and DeYoung]{W16-6204}
Mark Dredze, Nicholas Andrews, and Jay DeYoung.
\newblock Twitter at the grammys: A social media corpus for entity linking and
  disambiguation.
\newblock In \emph{Proceedings of The Fourth International Workshop on Natural
  Language Processing for Social Media}, pages 20--25. Association for
  Computational Linguistics, 2016.
\newblock \doi{10.18653/v1/W16-6204}.
\newblock URL \url{http://aclweb.org/anthology/W16-6204}.

\bibitem[Eshel et~al.(2017)Eshel, Cohen, Radinsky, Markovitch, Yamada, and
  Levy]{eshel-EtAl:2017:CoNLL}
Yotam Eshel, Noam Cohen, Kira Radinsky, Shaul Markovitch, Ikuya Yamada, and
  Omer Levy.
\newblock Named entity disambiguation for noisy text.
\newblock In \emph{Proceedings of the 21st Conference on Computational Natural
  Language Learning (CoNLL 2017)}, pages 58--68, Vancouver, Canada, August
  2017. Association for Computational Linguistics.
\newblock URL \url{http://aclweb.org/anthology/K17-1008}.

\bibitem[Fang and Chang(2014)]{Q14-1021}
Yuan Fang and Ming-Wei Chang.
\newblock Entity linking on microblogs with spatial and temporal signals.
\newblock \emph{Transactions of the Association for Computational Linguistics},
  2:\penalty0 259--272, 2014.
\newblock URL \url{http://aclweb.org/anthology/Q14-1021}.

\bibitem[Francis-Landau et~al.(2016)Francis-Landau, Durrett, and
  Klein]{francislandau-durrett-klein:2016:N16-1}
Matthew Francis-Landau, Greg Durrett, and Dan Klein.
\newblock Capturing semantic similarity for entity linking with convolutional
  neural networks.
\newblock In \emph{Proceedings of the 2016 Conference of the North American
  Chapter of the Association for Computational Linguistics: Human Language
  Technologies}, pages 1256--1261, San Diego, California, June 2016.
  Association for Computational Linguistics.
\newblock URL \url{http://www.aclweb.org/anthology/N16-1150}.

\bibitem[Galli et~al.(2015)Galli, Gurini, Gasparetti, Micarelli, and
  Sansonetti]{galli2015analysis}
Michele Galli, Davide~Feltoni Gurini, Fabio Gasparetti, Alessandro Micarelli,
  and Giuseppe Sansonetti.
\newblock Analysis of user-generated content for improving youtube video
  recommendation.
\newblock In \emph{RecSys Posters}, 2015.

\bibitem[Ganea and Hofmann(2017)]{D17-1277}
Octavian-Eugen Ganea and Thomas Hofmann.
\newblock Deep joint entity disambiguation with local neural attention.
\newblock In \emph{Proceedings of the 2017 Conference on Empirical Methods in
  Natural Language Processing}, pages 2619--2629. Association for Computational
  Linguistics, 2017.
\newblock \doi{10.18653/v1/D17-1277}.
\newblock URL \url{http://aclweb.org/anthology/D17-1277}.

\bibitem[Grover and Leskovec(2016)]{grover2016node2vec}
Aditya Grover and Jure Leskovec.
\newblock node2vec: Scalable feature learning for networks.
\newblock In \emph{Proceedings of the 22nd ACM SIGKDD international conference
  on Knowledge discovery and data mining}, pages 855--864. ACM, 2016.

\bibitem[Guo et~al.(2013{\natexlab{a}})Guo, Chang, and
  Kiciman]{guo-chang-kiciman:2013:NAACL-HLT}
Stephen Guo, Ming-Wei Chang, and Emre Kiciman.
\newblock To link or not to link? a study on end-to-end tweet entity linking.
\newblock In \emph{Proceedings of the 2013 Conference of the North American
  Chapter of the Association for Computational Linguistics: Human Language
  Technologies}, pages 1020--1030, Atlanta, Georgia, June 2013{\natexlab{a}}.
  Association for Computational Linguistics.
\newblock URL \url{http://www.aclweb.org/anthology/N13-1122}.

\bibitem[Guo et~al.(2013{\natexlab{b}})Guo, Qin, Liu, and
  Li]{guo-EtAl:2013:EMNLP}
Yuhang Guo, Bing Qin, Ting Liu, and Sheng Li.
\newblock Microblog entity linking by leveraging extra posts.
\newblock In \emph{Proceedings of the 2013 Conference on Empirical Methods in
  Natural Language Processing}, pages 863--868, Seattle, Washington, USA,
  October 2013{\natexlab{b}}. Association for Computational Linguistics.
\newblock URL \url{http://www.aclweb.org/anthology/D13-1085}.

\bibitem[Gupta et~al.(2017)Gupta, Singh, and
  Roth]{gupta-singh-roth:2017:EMNLP2017}
Nitish Gupta, Sameer Singh, and Dan Roth.
\newblock Entity linking via joint encoding of types, descriptions, and
  context.
\newblock In \emph{Proceedings of the 2017 Conference on Empirical Methods in
  Natural Language Processing}, pages 2681--2690, Copenhagen, Denmark,
  September 2017. Association for Computational Linguistics.
\newblock URL \url{https://www.aclweb.org/anthology/D17-1284}.

\bibitem[Hoffart et~al.(2011{\natexlab{a}})Hoffart, Yosef, Bordino,
  F{\"u}rstenau, Pinkal, Spaniol, Taneva, Thater, and Weikum]{D11-1072}
Johannes Hoffart, Mohamed~Amir Yosef, Ilaria Bordino, Hagen F{\"u}rstenau,
  Manfred Pinkal, Marc Spaniol, Bilyana Taneva, Stefan Thater, and Gerhard
  Weikum.
\newblock Robust disambiguation of named entities in text.
\newblock In \emph{Proceedings of the 2011 Conference on Empirical Methods in
  Natural Language Processing}, pages 782--792. Association for Computational
  Linguistics, 2011{\natexlab{a}}.
\newblock URL \url{http://aclweb.org/anthology/D11-1072}.

\bibitem[Hoffart et~al.(2011{\natexlab{b}})Hoffart, Yosef, Bordino,
  F\"{u}rstenau, Pinkal, Spaniol, Taneva, Thater, and
  Weikum]{hoffart-EtAl:2011:EMNLP}
Johannes Hoffart, Mohamed~Amir Yosef, Ilaria Bordino, Hagen F\"{u}rstenau,
  Manfred Pinkal, Marc Spaniol, Bilyana Taneva, Stefan Thater, and Gerhard
  Weikum.
\newblock Robust disambiguation of named entities in text.
\newblock In \emph{Proceedings of the 2011 Conference on Empirical Methods in
  Natural Language Processing}, pages 782--792, Edinburgh, Scotland, UK., July
  2011{\natexlab{b}}. Association for Computational Linguistics.
\newblock URL \url{http://www.aclweb.org/anthology/D11-1072}.

\bibitem[Hua et~al.(2015)Hua, Zheng, and Zhou]{hua2015microblog}
Wen Hua, Kai Zheng, and Xiaofang Zhou.
\newblock Microblog entity linking with social temporal context.
\newblock In \emph{Proceedings of the 2015 ACM SIGMOD International Conference
  on Management of Data}, pages 1761--1775. ACM, 2015.

\bibitem[Huang et~al.(2014)Huang, Cao, Huang, Ji, and
  Lin]{huang-EtAl:2014:P14-11}
Hongzhao Huang, Yunbo Cao, Xiaojiang Huang, Heng Ji, and Chin-Yew Lin.
\newblock Collective tweet wikification based on semi-supervised graph
  regularization.
\newblock In \emph{Proceedings of the 52nd Annual Meeting of the Association
  for Computational Linguistics (Volume 1: Long Papers)}, pages 380--390,
  Baltimore, Maryland, June 2014. Association for Computational Linguistics.
\newblock URL \url{http://www.aclweb.org/anthology/P14-1036}.

\bibitem[J{\"a}rvelin and Kek{\"a}l{\"a}inen(2002)]{jarvelin2002cumulated}
Kalervo J{\"a}rvelin and Jaana Kek{\"a}l{\"a}inen.
\newblock Cumulated gain-based evaluation of ir techniques.
\newblock \emph{ACM Transactions on Information Systems (TOIS)}, 20\penalty0
  (4):\penalty0 422--446, 2002.

\bibitem[Ji et~al.(2010)Ji, Grishman, and Dang]{ji2010overview}
Heng Ji, Ralph Grishman, and Hoa~Trang Dang.
\newblock Overview of the tac 2010 knowledge base population track.
\newblock 2010.

\bibitem[Ji et~al.(2016)Ji, Nothman, Dang, and Hub]{ji2016overview}
Heng Ji, Joel Nothman, H~Trang Dang, and Sydney~Informatics Hub.
\newblock Overview of tac-kbp2016 tri-lingual edl and its impact on end-to-end
  cold-start kbp.
\newblock \emph{Proceedings of TAC}, 2016.

\bibitem[Kataria et~al.(2011)Kataria, Kumar, Rastogi, Sen, and
  Sengamedu]{kataria2011entity}
Saurabh~S Kataria, Krishnan~S Kumar, Rajeev~R Rastogi, Prithviraj Sen, and
  Srinivasan~H Sengamedu.
\newblock Entity disambiguation with hierarchical topic models.
\newblock In \emph{Proceedings of the 17th ACM SIGKDD international conference
  on Knowledge discovery and data mining}, pages 1037--1045. ACM, 2011.

\bibitem[Kazama and Torisawa(2007)]{kazama-torisawa:2007:EMNLP-CoNLL20072}
Jun'ichi Kazama and Kentaro Torisawa.
\newblock Exploiting {Wikipedia} as external knowledge for named entity
  recognition.
\newblock In \emph{Proceedings of the 2007 Joint Conference on Empirical
  Methods in Natural Language Processing and Computational Natural Language
  Learning (EMNLP-CoNLL)}, pages 698--707, Prague, Czech Republic, June 2007.
  Association for Computational Linguistics.
\newblock URL \url{http://www.aclweb.org/anthology/D/D07/D07-1073}.

\bibitem[Kingma and Ba(2015)]{kingma2014adam}
Diederik~P Kingma and Jimmy Ba.
\newblock Adam: A method for stochastic optimization.
\newblock In \emph{Proceedings of the International Conference on Learning
  Representations (ICLR)}, 2015.

\bibitem[Lau et~al.(2012)Lau, Collier, and
  Baldwin]{lau-collier-baldwin:2012:PAPERS}
Jey~Han Lau, Nigel Collier, and Timothy Baldwin.
\newblock On-line trend analysis with topic models: \#twitter trends detection
  topic model online.
\newblock In \emph{Proceedings of COLING 2012}, pages 1519--1534, Mumbai,
  India, December 2012. The COLING 2012 Organizing Committee.
\newblock URL \url{http://www.aclweb.org/anthology/C12-1093}.

\bibitem[Le and Titov(2018)]{P18-1148}
Phong Le and Ivan Titov.
\newblock Improving entity linking by modeling latent relations between
  mentions.
\newblock In \emph{Proceedings of the 56th Annual Meeting of the Association
  for Computational Linguistics (Volume 1: Long Papers)}, pages 1595--1604.
  Association for Computational Linguistics, 2018.
\newblock URL \url{http://aclweb.org/anthology/P18-1148}.

\bibitem[Liu et~al.(2013)Liu, Li, Wu, Zhou, Wei, and
  Lu]{liu-EtAl:2013:ACL20133}
Xiaohua Liu, Yitong Li, Haocheng Wu, Ming Zhou, Furu Wei, and Yi~Lu.
\newblock Entity linking for tweets.
\newblock In \emph{Proceedings of the 51st Annual Meeting of the Association
  for Computational Linguistics (Volume 1: Long Papers)}, pages 1304--1311,
  Sofia, Bulgaria, August 2013. Association for Computational Linguistics.
\newblock URL \url{http://www.aclweb.org/anthology/P13-1128}.

\bibitem[Luong et~al.(2015)Luong, Pham, and
  Manning]{luong-pham-manning:2015:EMNLP}
Thang Luong, Hieu Pham, and Christopher~D. Manning.
\newblock Effective approaches to attention-based neural machine translation.
\newblock In \emph{Proceedings of the 2015 Conference on Empirical Methods in
  Natural Language Processing}, pages 1412--1421, Lisbon, Portugal, September
  2015. Association for Computational Linguistics.
\newblock URL \url{http://aclweb.org/anthology/D15-1166}.

\bibitem[Messenger and Whittle(2011)]{messenger2011recommendations}
Andrew Messenger and Jon Whittle.
\newblock Recommendations based on user-generated comments in social media.
\newblock In \emph{Privacy, Security, Risk and Trust (PASSAT) and 2011 IEEE
  Third Inernational Conference on Social Computing (SocialCom), 2011 IEEE
  Third International Conference on}, pages 505--508. IEEE, 2011.

\bibitem[Mikolov et~al.(2013)Mikolov, Chen, Corrado, and
  Dean]{mikolov2013efficient}
Tomas Mikolov, Kai Chen, Greg Corrado, and Jeffrey Dean.
\newblock Efficient estimation of word representations in vector space.
\newblock In \emph{Proceedings of the International Conference on Learning
  Representations (ICLR)}, 2013.

\bibitem[Mintz et~al.(2009)Mintz, Bills, Snow, and Jurafsky]{P09-1113}
Mike Mintz, Steven Bills, Rion Snow, and Daniel Jurafsky.
\newblock Distant supervision for relation extraction without labeled data.
\newblock In \emph{Proceedings of the Joint Conference of the 47th Annual
  Meeting of the ACL and the 4th International Joint Conference on Natural
  Language Processing of the AFNLP}, pages 1003--1011. Association for
  Computational Linguistics, 2009.
\newblock URL \url{http://aclweb.org/anthology/P09-1113}.

\bibitem[Moon et~al.(2018)Moon, Neves, and Carvalho]{N18-1078}
Seungwhan Moon, Leonardo Neves, and Vitor Carvalho.
\newblock Multimodal named entity recognition for short social media posts.
\newblock In \emph{Proceedings of the 2018 Conference of the North American
  Chapter of the Association for Computational Linguistics: Human Language
  Technologies, Volume 1 (Long Papers)}, pages 852--860. Association for
  Computational Linguistics, 2018.
\newblock \doi{10.18653/v1/N18-1078}.
\newblock URL \url{http://aclweb.org/anthology/N18-1078}.

\bibitem[Noreen(1989)]{noreen1989computer}
Eric~W Noreen.
\newblock \emph{Computer-intensive methods for testing hypotheses}.
\newblock Wiley New York, 1989.

\bibitem[O'Connor et~al.(2010)O'Connor, Balasubramanyan, Routledge, Smith,
  et~al.]{o2010tweets}
Brendan O'Connor, Ramnath Balasubramanyan, Bryan~R Routledge, Noah~A Smith,
  et~al.
\newblock From tweets to polls: Linking text sentiment to public opinion time
  series.
\newblock \emph{Icwsm}, 11\penalty0 (122-129):\penalty0 1--2, 2010.

\bibitem[Ratinov et~al.(2011)Ratinov, Roth, Downey, and
  Anderson]{ratinov-EtAl:2011:ACL-HLT2011}
Lev Ratinov, Dan Roth, Doug Downey, and Mike Anderson.
\newblock Local and global algorithms for disambiguation to wikipedia.
\newblock In \emph{Proceedings of the 49th Annual Meeting of the Association
  for Computational Linguistics: Human Language Technologies}, pages
  1375--1384, Portland, Oregon, USA, June 2011. Association for Computational
  Linguistics.
\newblock URL \url{http://www.aclweb.org/anthology/P11-1138}.

\bibitem[Shen et~al.(2013)Shen, Wang, Luo, and Wang]{shen2013linking}
Wei Shen, Jianyong Wang, Ping Luo, and Min Wang.
\newblock Linking named entities in tweets with knowledge base via user
  interest modeling.
\newblock In \emph{Proceedings of the 19th ACM SIGKDD international conference
  on Knowledge discovery and data mining}, pages 68--76. ACM, 2013.

\bibitem[Shen et~al.(2015)Shen, Wang, and Han]{shen2015entity}
Wei Shen, Jianyong Wang, and Jiawei Han.
\newblock Entity linking with a knowledge base: Issues, techniques, and
  solutions.
\newblock \emph{IEEE Transactions on Knowledge and Data Engineering},
  27\penalty0 (2):\penalty0 443--460, 2015.

\bibitem[Sun et~al.(2015)Sun, Lin, Tang, Yang, Ji, and Wang]{sun2015modeling}
Yaming Sun, Lei Lin, Duyu Tang, Nan Yang, Zhenzhou Ji, and Xiaolong Wang.
\newblock Modeling mention, context and entity with neural networks for entity
  disambiguation.
\newblock In \emph{IJCAI}, pages 1333--1339, 2015.

\bibitem[Voorhees et~al.(1999)]{voorhees1999trec}
Ellen~M Voorhees et~al.
\newblock The trec-8 question answering track report.
\newblock In \emph{Trec}, volume~99, pages 77--82, 1999.

\bibitem[Yamada et~al.(2016)Yamada, Shindo, Takeda, and Takefuji]{K16-1025}
Ikuya Yamada, Hiroyuki Shindo, Hideaki Takeda, and Yoshiyasu Takefuji.
\newblock Joint learning of the embedding of words and entities for named
  entity disambiguation.
\newblock In \emph{Proceedings of The 20th SIGNLL Conference on Computational
  Natural Language Learning}, pages 250--259. Association for Computational
  Linguistics, 2016.
\newblock \doi{10.18653/v1/K16-1025}.
\newblock URL \url{http://aclweb.org/anthology/K16-1025}.

\bibitem[Yang and Chang(2015)]{yang-chang:2015:ACL-IJCNLP}
Yi~Yang and Ming-Wei Chang.
\newblock S-mart: Novel tree-based structured learning algorithms applied to
  tweet entity linking.
\newblock In \emph{Proceedings of the 53rd Annual Meeting of the Association
  for Computational Linguistics and the 7th International Joint Conference on
  Natural Language Processing (Volume 1: Long Papers)}, pages 504--513,
  Beijing, China, July 2015. Association for Computational Linguistics.
\newblock URL \url{http://www.aclweb.org/anthology/P15-1049}.

\bibitem[Yang et~al.(2016)Yang, Chang, and
  Eisenstein]{yang-chang-eisenstein:2016:EMNLP2016}
Yi~Yang, Ming-Wei Chang, and Jacob Eisenstein.
\newblock Toward socially-infused information extraction: Embedding authors,
  mentions, and entities.
\newblock In \emph{Proceedings of the 2016 Conference on Empirical Methods in
  Natural Language Processing}, pages 1452--1461, Austin, Texas, November 2016.
  Association for Computational Linguistics.
\newblock URL \url{https://aclweb.org/anthology/D16-1152}.

\bibitem[Zhao et~al.(2014)Zhao, Guo, He, Jiang, Wu, and Li]{zhao2014we}
Xin~Wayne Zhao, Yanwei Guo, Yulan He, Han Jiang, Yuexin Wu, and Xiaoming Li.
\newblock We know what you want to buy: a demographic-based system for product
  recommendation on microblogs.
\newblock In \emph{Proceedings of the 20th ACM SIGKDD international conference
  on Knowledge discovery and data mining}, pages 1935--1944. ACM, 2014.

\bibitem[Zheng et~al.(2010)Zheng, Li, Huang, and Zhu]{N10-1072}
Zhicheng Zheng, Fangtao Li, Minlie Huang, and Xiaoyan Zhu.
\newblock Learning to link entities with knowledge base.
\newblock In \emph{Human Language Technologies: The 2010 Annual Conference of
  the North American Chapter of the Association for Computational Linguistics},
  pages 483--491. Association for Computational Linguistics, 2010.
\newblock URL \url{http://aclweb.org/anthology/N10-1072}.

\bibitem[Zwicklbauer et~al.(2016)Zwicklbauer, Seifert, and
  Granitzer]{zwicklbauer2016robust}
Stefan Zwicklbauer, Christin Seifert, and Michael Granitzer.
\newblock Robust and collective entity disambiguation through semantic
  embeddings.
\newblock In \emph{Proceedings of the 39th International ACM SIGIR conference
  on Research and Development in Information Retrieval}, pages 425--434. ACM,
  2016.

\end{thebibliography}
\bibliographystyle{plainnat}

\end{document}